\PassOptionsToPackage{dvipsnames}{xcolor}
 
\documentclass{article}

\usepackage[utf8]{inputenc} 
\usepackage[T1]{fontenc}    
\usepackage{hyperref}       
\usepackage{url}            
\usepackage{booktabs}       
\usepackage{amsfonts}       
\usepackage{nicefrac}       
\usepackage{microtype}      
\usepackage{lipsum}		
\usepackage{graphicx}
\usepackage{float}
\usepackage{doi}
\usepackage[normalem]{ulem}

\usepackage{tabularx}
\usepackage{color}
\usepackage{colortbl}
\usepackage{lineno}
\usepackage{xcolor}

\usepackage{soul}
\usepackage{arxiv}
\usepackage{url}
\newcommand\myshade{85}
\colorlet{mylinkcolor}{RoyalBlue}
\colorlet{mycitecolor}{violet}
\colorlet{myurlcolor}{YellowOrange}

\hypersetup{
  linkcolor  = mylinkcolor!\myshade!black,
  citecolor  = mycitecolor!\myshade!black,
  urlcolor   = myurlcolor!\myshade!black,
  colorlinks = true,
}

\usepackage{times}
\usepackage{latexsym}

\usepackage[T1]{fontenc}

\usepackage[utf8]{inputenc}

\usepackage{microtype}

\usepackage{booktabs}
\usepackage{graphicx}
\usepackage{array}
\usepackage{multirow}
\usepackage{longtable}
\usepackage{CJKutf8}
\usepackage{float}
\usepackage{microtype}
\usepackage{multirow}
\usepackage{booktabs}
\usepackage{graphicx}
\usepackage{amssymb}
\usepackage{url}
\usepackage{amsmath}
\usepackage{caption}
\usepackage[ruled,vlined,linesnumbered]{algorithm2e}
\usepackage{subfigure}
\usepackage{natbib}
\usepackage{makecell}
\usepackage{ulem}
\usepackage{fancyhdr}
\usepackage{tikz}
\usetikzlibrary{shapes.geometric}
\newcommand\blfootnote[1]{%
  \begingroup
  \renewcommand\thefootnote{}\footnote{#1}%
  \addtocounter{footnote}{-1}%
  \endgroup
}

\title{Mixup-Augmented Meta-Learning for \\Sample-Efficient Fine-Tuning of Protein Simulators}

\author{ 
    \small
    Jingbang Chen\textsuperscript{1*},\space
    Yian Wang\textsuperscript{2*},\space
    Xingwei Qu \textsuperscript{3},\space
    Shuangjia Zheng  \textsuperscript{4},\space
    Yaodong Yang \textsuperscript{2\dag},\space 
    Hao Dong \textsuperscript{2\dag},\space
    Jie Fu\textsuperscript{3\dag}\\ 
{\small 
\textsuperscript{1} University of California, Los Angeles
}\vspace{-0.1mm} \\ 
{\small 
\textsuperscript{2} Peking University
}\vspace{-0.1mm} \\
{\small 
\textsuperscript{3} Hong Kong University of Science and Technology
}\vspace{-0.1mm} \\
{\small 
\textsuperscript{4} Shanghai Jiao Tong University
}\vspace{-0.1mm}  \\
\texttt{\small
jingbangchen@g.ucla.edu
}\vspace{-0.1mm} \\
\texttt{\small
\{yianwang, 
	yaodong.yang, 
	hao.dong\}@pku.edu.cn
}\vspace{-0.1mm} \\
\texttt{\small
xingwei.qu@tum.de
}\vspace{-0.1mm} \\
\texttt{\small
shuangjia.zheng@sjtu.edu.cn
}\vspace{-0.1mm} \\
\texttt{\small
jiefu@ust.hk
}\vspace{-0.3cm} \\
}

\begin{document}
\blfootnote{* Equal contribution.}
\blfootnote{\dag Corresponding authors, equal advising.}
\maketitle

\begin{abstract}
Molecular dynamics simulations have emerged as a fundamental instrument for studying biomolecules. At the same time, it is desirable to perform simulations of a collection of particles under various conditions in which the molecules can fluctuate. In this paper, we explore and adapt the soft prompt-based learning method to molecular dynamics tasks. Our model can remarkably generalize to unseen and out-of-distribution scenarios with limited training data. While our work focuses on temperature as a test case, the versatility of our approach allows for efficient simulation through any continuous dynamic conditions, such as pressure and volumes.
Our framework has two stages: 1) Pre-trains with data mixing technique, augments molecular structure data and temperature prompts, then applies a curriculum learning method by increasing the ratio of them smoothly. 
2) Meta-learning-based fine-tuning framework improves sample-efficiency of fine-tuning process and gives the soft prompt-tuning better initialization points. Comprehensive experiments reveal that our framework excels in accuracy for in-domain data and demonstrates strong generalization capabilities for unseen and out-of-distribution samples. Code is provided at: \href{https://github.com/Jingbang-Chen/mixup-meta-protein-simulators}{https://github.com/Jingbang-Chen/mixup-meta-protein-simulators.}
\end{abstract}

\section{Introduction}
Graph representation deep learning has reached huge success in protein simulations and molecular dynamics with high efficiency, where proteins can be trained according to their 3D structures. For instance, \cite{zhang2022protein} pre-train protein sequence by multiview contrastive learning. DeepRank-GNN \citep{reau2023deeprank} utilize pre-defined GNN architecture to learn the problem-specific protein-protein interaction patterns. Additionally, generative networks with equivariant properties have successfully predicted equilibrium distributions, enabling protein conformation sampling \citep{zheng2023towards}. 
These suggest that graph representation learning is a good marriage of predicting various molecule structures.

Recently, most papers have focused on pre-training protein language models, using a substantial number of unlabeled amino acid sequences, followed by fine-tuning with labeled data in downstream tasks \citep{madani2020progen, somnath2021multi}.  However, there is limited research on sample-efficient protein simulation under different conditions, such as varying temperatures. In scenerios where labeled and unlabeled data are limited, these large-scale pre-training models are difficult to train.

Motivated by this development, we propose a soft prompt-based method to incorporate different prompts into molecule prediction tasks, aiming to predict the structure of proteins at different conditions. Here we choose temperature as a condition as it plays a pivotal role in determining the energy of proteins, which in turn influences their structure. We show that our method can generalize to predict molecular dynamics under different temperatures, which are unseen and out-of-distribution in training data. 
We view different temperatures as different prompts and process the temperatures and structure information using two different encoders, a prompt encoder, and a prompt-agnostic encoder. To accurately learn the latent space of prompt vectors, which are mostly discrete tokens, recent works \citep{qin2021learning, li2021prefix} in language modeling have introduced soft prompts that utilize learnable prompt vectors. However, traditional soft prompts pose challenges in learning and are sensitive to initialization points when dealing with limited data \citep{liu2023pre}.  Additionally, in certain biological tasks, the training data is collected from the real world, making it impractical to gather sufficient data under various environmental conditions. To this end, we propose to use a data augmentation skill to shape the latent space's distribution. Our method draws inspiration from the widely used mixing method in the computer vision domain \citep{lemley2017smart, devries2017dataset}, which masks parts of the image and replaces with other images thereby enhancing training robustness. We further extend the idea of mixup methods to the field of prompt-based pre-training by using two mixup networks separately in the domain of temperatures and structures. 
Furthermore, we employ curriculum learning to make the whole training much easier. 
In this way, we force the latent space to be continuous rather than discrete so that we can not only enable a smooth learning space of prompts, but also enhance the whole pre-training procedure, which helps our network generalize to unseen and out-of-distribution prompts very well.

To achieve a rapid and efficient fine-tuning process, we employ meta-training to obtain well-initialized starting points for different prompts, as the results can vary significantly under randomly initialized parameters. We use an optimization-based meta-learning algorithm inspired by MAML \citep{finn2017modelagnostic} in the second-stage pre-training to find the best initial parameters.

We conduct comprehensive experiments on protein structure prediction to evaluate the effects of each component. Numerical results show that our two-stage pre-training methods can help the network predict the positions of atoms under unseen and even out-of-distribution temperatures. Indeed, this capability has the potential to greatly expand the scope of molecular structure prediction, especially when dealing with limited data and varying conditions. This advancement can lead to significant advancements in various fields, including drug discovery and protein engineering, where predicting molecule structures accurately under different conditions is crucial.

In conclusion, our key contributions are as follows:
\begin{itemize} 
    \item 
    We propose a protein simulation network under different temperatures via prompt-based tuning methods, which includes two-training stages conducted in a series manner. 
    \item 
    The first stage combines prompt-tuning methods with data augmentation methods, which effectively shapes the latent space of prompt vectors despite the scarcity of data.
    \item 
    We combine meta-learning and mixup to obtain an appropriate initial parameter of the prompt vector before fine-tuning, which improves the data efficiency, accelerates the fine-tuning process, and generalizes to out-of-distribution cases better.
\end{itemize}

\section{Related Work}
\label{related}
\begin{itemize} 

\item
    \textbf{Equivariant Graph Neural Networks (EGNN) in Molecule Simulation} To achieve equivariance, various approaches have been proposed in the literature. \cite{kohler2019equivariant} explore the concept of equivariance in the context of graph neural networks and demonstrated its potential for improving predictive performance in molecular tasks. \cite{thomas2018tensor} introduce the use of tensor network-based equivariant architectures, showcasing their effectiveness in processing molecular data while maintaining symmetry under relevant transformations. \cite{weiler20183d} extend the concept of equivariance to 3D structures, allowing the model to handle spatial rotations and translations for improved accuracy in 3D graph data processing.

    In the domain of protein simulation, graph neural networks (GNNs) with equivariant properties have emerged as a powerful approach to capture the symmetries and physical constraints present in protein structures \citep{jing2020learning, aykent2022gbpnet, chen20233d}. Recently,  EGNNs \citep{satorras2022en} generalize the equivariant GNN framework into N dimension, extracting both invariant and equivariant embeddings, which is also widely adopted in learning protein structures \citep{zhang2022protein, trippe2022diffusion}.

    \item 
    \textbf{Mixing Augmentation} Data mixing augmentation has boosted robust performance in computer vision tasks. At first, data augmentation in computer vision employed the masking teniques to some part of the image, adding noise into the image which makes the network much robust. \cite{devries2017improved}
     remove random patches section from the image during training, while \cite{zhong2020random} erase random rectangles of input images instead of squares. This approach further enhances the robustness of the model and has shown promising results in various computer vision tasks. After that, a series of mixing methods emerged, including pixel-based mixing and patch-based mixing. The foundation one called Mixup \citep{zhang2018mixup} linearly combines corresponding data pairs and labels to augment samples. Manifold Mixup \citep{verma2019manifold} extends mixup to a higher-dimensional latent space for better performance.  CutMix \citep{yun2019cutmix} designs a patch replacement strategies replaces part of patches from other data samples. Other works like PuzzleMix \citep{kim2020puzzle}, Co-Mixup \citep{kim2021co} and Automix \citep{liu2022automix} are based on an iterative optimization process, to find the optimal mixing behavior given a trainable mixing network.

    Beyond computer vision tasks, mixup has found applications in other domains as well. TextMix \citep{yoon2021ssmix} applies mixup in natural language processing by interpolating input text and their corresponding labels, which enhances the generalization of text-based models and has demonstrated improved performance on various NLP tasks. Audio Mixup \citep{min202210} extends the mixup concept to audio data. By linearly combining audio samples and their labels, Audio Mixup enhances the performance of speech and audio processing models, showcasing the versatility of mixup as a data augmentation technique across different domains. The success and adaptability of mixup and its variants have established them as valuable tools in the machine learning practitioner's arsenal. These techniques not only improve the model's predictive capabilities but also contribute to a more comprehensive understanding of generalization and regularization in deep learning.

    
    \item 
    \textbf{Meta Learning} Meta-learning is an important algorithm used in few-shot learning tasks. It can be classified into metric/similarity-based methods and optimization-based methods. The former domain learns a latent space that according to task-specific similarity \citep{koch2015siamese, vinyals2016matching}. The latter  utilizes meta-learning in the optimization process, receiving inner loop gradients and updating other parameters. Examples of optimization-based meta-learning algorithms include Reptile \citep{nichol2018reptile} and Model-Agnostic Meta-Learning (MAML) \citep{finn2017modelagnostic}, which adapt model parameters across tasks to enable fast adaptation to new tasks with limited data.

\end{itemize}

\section{Preliminaries}

\subsection{Task Definition}
\begin{figure*}[!ht]
    \centering
    \includegraphics[scale=0.4]{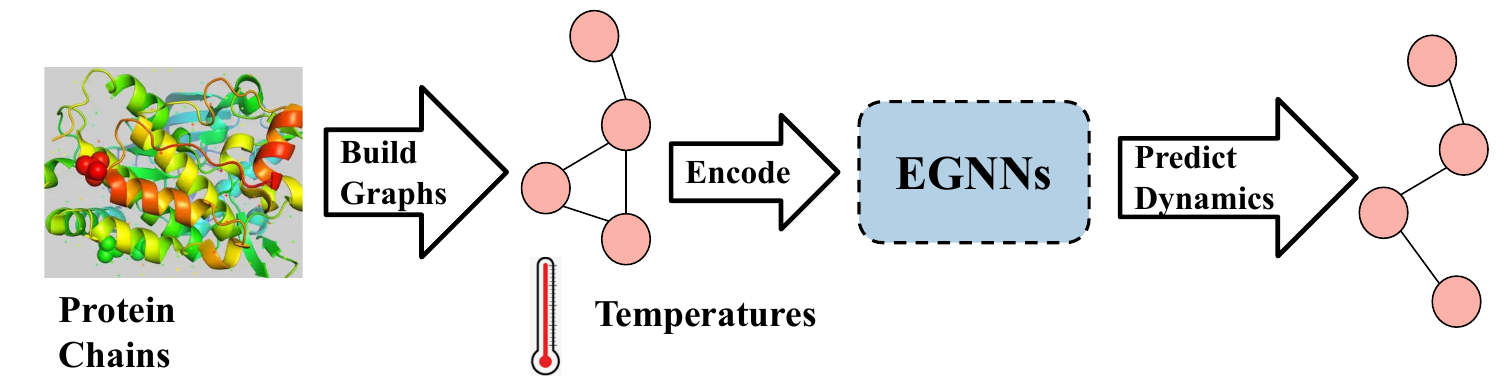}
    \caption{Task description.}
    \label{fig:task}
\end{figure*}

As shown in figure \ref{fig:task}, the task of molecule structure prediction is to predict the molecule structure under different temperatures. The molecule structure inputs include atom positions $x = (x_1,...,x_N) \in \mathbb{R} ^{N \times 3}$, atom embeddings $h = (h_1,...,h_N) \in \mathbb{R}^{N \times d}$. Our goal is to predict positions $x$ under different temperatures $T$. And we adopt the mean squared error (MSE) as the metric to calculate the loss between the predicted positions $x$ and ground-truth ones $x^{GT}$ as $L_{MSE}(x, x^{GT}) = \frac{1}{N}\sum\limits_{n=1}^N(x_i-x_i^{GT})^2$.

\subsection{Equivariance}

Formally, a function $\mathcal{F}: \mathcal{X} \rightarrow \mathcal{Y}$ is defiend as equivariant w.r.t the action of a group $G$ if:
    \begin{equation}
        \mathcal{F} \circ \mathcal{T}_g(x) = \mathcal{S}_g \circ \mathcal{F}(x), \forall g \in G,
    \end{equation}
where $S_g, T_g$ are transformations for a group element $g \in G$, operating on the vector spaces $\mathcal{X}$ and $\mathcal{Y}$ respectively. In this work, we consider the SE(3) group, i.e., the group of rotation and translation in 3D space. In this setting, the transformations $\mathcal{T}_g$ and $\mathcal{S}_g$ can be represented by a translation $\bf{t}$ and orthogonal matrix rotation $\bf{R}$.

In our molecule conformation prediction task, we input the node embeddings $\bf{h}^t$ and coordinate features $\bf{x}^t$ into the equivariant graph neural networks (EGNNs) \citep{satorras2022en}. The node embeddings are SE(3)-invariant while the coordinates are equivariant, expressed as $\bf{Rx} + \bf{t} = (\bf{Rx_1+t}, ..., \bf{Rx_N + t}) $. The network outputs the node embeddings $\bf{h}^{t+1}$ and coordinates feature $\bf{x}^{t+1}$ of next time after L EGCL layers transformation.

\section{Methodology}
\label{Methodology}
In this session, we elaborate on our proposed framework. We first introduce the model architecture of our backbone network that leverages temperature prompts to encode MD trajectories. Next, we introduce our mixup policy which enhances the training difficulty and ensures a smoother and more continuous latent space.  By incrementally increasing the training difficulty through this mixup policy, we can shape the latent space to be more smooth and continuous. Subsequently, we illustrate the meta training process, giving different prompts an optimal initialization point. Finally, we give a brief demonstration of our testing process. The notation used in this session is summarized in table \ref{tab:notation}.\\\

\begin{table}[!ht]
    \centering
    \caption{A reference table of notations used in the methodology session}
    \label{tab:notation}
    \renewcommand\arraystretch{1.1}
    \begin{tabularx}{\columnwidth}{cX}
        \hline
        Symbol & Definition \\
        \hline \\
        \multicolumn{2}{c}
        {\underline{\emph{Hyperparameters in first-stage pre-training:}}} \\
        $E_{pre}$ & The number of pre-training epochs before using mixup networks \\
        $N_{period}$ & The ratio of real training data to the synthesized data used in training the backbone network(other parts of the network except mixup networks) \\
        $E_{period}$ & The number of epochs which halves $N_{period}$ in curriculum learning \\
        $N_{min}$ & The minimum of $N_{period}$ \\
        \\
        \multicolumn{2}{c}{\underline{\emph{Model parameters and variables in first-stage pre-training:}}} \\
        $f_{\phi_P}$ & PromptMix Network \\
        $P_i$ & Prompt \\
        $P_m$ & Mixed Prompt \\
        $f_{\phi_S}$ & StructMix Network \\
        $S_i$ & Vector of structure feature \\
        $S_m$ & Mixed vector of structure feature \\
        $x$ & Predicted positions \\
        $x^{GT}$ & Ground-truth of predicted positions \\
        $x_m$ & Predicted positions of mixed data \\
        $L$ & Loss calculated by real data \\
        $L_m$ & Loss calculated by mixed data \\
        $L_m^\theta$ & Loss calculated by mixed data and backpropagated to the backbone network (other parts of the network except mixup networks) \\
        $L_m^\phi$ & Loss calculated by mixed data and backpropagated to the mixup networks \\
        $L_{first}$ & Total loss of first-stage pre-training
        \\
        \multicolumn{2}{c}{\underline{\emph{Model parameters and variables in prompt meta-training:}}} \\
        $D_{sup}$ & Data of support set \\
        $D_{que}$ & Data of query set \\
        $\theta_{P_i}$ & Parameters of the prompt encoder of prompt i \\
        $\theta_O$ & Parameters of the prompt-agnostic encoder and the pre-trained network (EGNNs) \\
        $\phi_{P_i}$ & Parameters of the PromptMix of prompt i \\
        \hline
    \end{tabularx}
\end{table}

\subsection{Backbone Network Architecture}
\label{baseline}

\begin{figure*}[!h]
    \centering
    \includegraphics[scale=0.8]{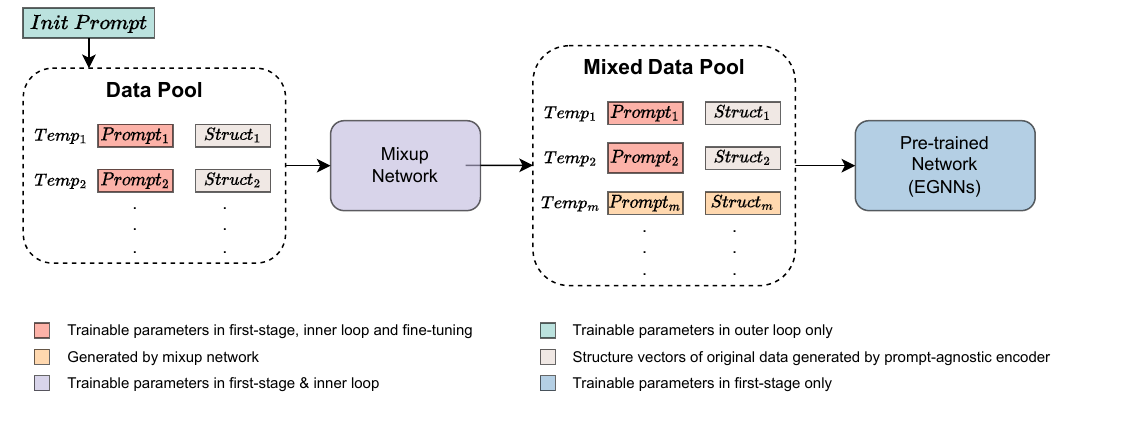}
    \caption{Overall model architecture. The prompt-agnostic encoder is omitted in this framework, which is trainable in the first stage only. All other parameters are color-coded according to their respective training stages.}
    \label{fig:overall}
\end{figure*}

Our schema consists of a two-stage pre-training process as shown in figure \ref{fig:overall}. Our backbone network architecture includes a prompt-agnostic encoder, prompt encoder, pre-training network, and mixup networks. We denote the parameters of the mixup networks as $\phi$ and the parameters of the other components as $\theta$. We adopt EGNNs as our pre-training network as they satisfy the equivariant property and effectively learn to simulate the protein structure of the next timestep given different prompts. \\

\textbf{Encoder} \ 
Our prompt-agnostic encoder is designed to encode structure information that is invariant to different prompts. We denote the output of this net as $S$. Its primary function is to extract high-quality features from the molecule structure and facilitate the training of the StructMix net. This encoder consists of a single EGNN layer. On the other hand, the prompt encoder is just a multi-linear perceptron network, which we will refer to as "prompt" in the following sections.\\

\subsection{First-Stage Pre-training}  
\label{first stage}
\begin{figure*}[h]
    \centering
    \includegraphics[scale=0.8]{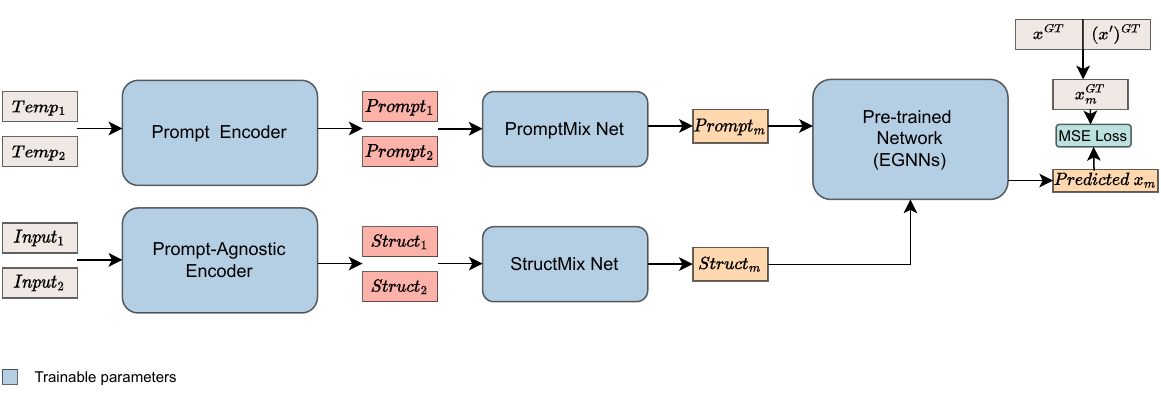}
    \caption{Illustration of first-stage pre-training procedures. }
    \label{fig:automix}
\end{figure*}

\begin{algorithm}[H]
\caption{First-stage Pre-training}
\label{alg:first-stage}
\KwIn{$E_{pre},E_{period},N_{period},N_{min}$: hyperparameters}
\While{not done}  
{

\For{ Epoch $E=1,2,...,E_{pre}$ }
{Update $\theta$ an epoch via loss $L$\;} 
\For{ Epoch $E=E_{pre}+1,...$ }
{\If{$(E-E_{pre}) \% E_{period} == 0 $ and $N_{period}>N_{min}$} 
{$N_{period} /= 2$ }

Filling data pool of structure features\\
Update $\phi$ for a epoch using $L_m^\phi$\\
\For{ Training Data $N = 1,...,n$}
{ Update $\theta$ a step via loss $L$\\
\If{$N \% N_{period} == 0 $ } 
{Update $\theta$ a step via loss $L_{m}^\theta$ }

}
}

}
\end{algorithm}
In the first-stage of pre-training, we employ the concept of curriculum learning \citep{bengio2009curriculum} with mixup networks. This approach involves prioritizing the learning process by initially focusing on a subset of simple training examples and gradually incorporating more challenging samples. In our specific settings, we train the pre-trained EGNN network for $E_{pre}$ epochs. Subsequently, we integrate synthesizing training data and labels generated by mixup networks to our training every $N_{period}$ real training samples, with $N_{period}$ being halved every $E_{period}$ epochs. This process will hold until $N_{period} \le N_{min}$.  The whole procedure is outlined in algorithm \ref{alg:first-stage}. In this way, we can train the whole network using the original data first, followed by incorporating some virtual mixed data to boost its robustness, instead of introducing the mixup strategy at the beginning of training, which might hinder the model to learn knowledge from the real data. The overall structure is shown in figure \ref{fig:automix}.\\

\textbf{Mixup Networks}
Our mixup networks consist of two components: StructMix $f_{\phi_{S}}$ and PromptMix $f_{\phi_{P}}$. StructMix $f_{\phi_{S}}$ is responsible for generating a mixed prompt-invariant structure feature vector, while PromptMix $f_{\phi_{P}}$ produces a mixed prompt vector. These two networks can independently make new prompts and structure features, thereby augmenting our training database.

While incorporating the training of mixup networks, we first use the prompt-agnostic encoder and prompt encoder to get a data pool consisting of both prompt vectors and struct vectors of all training data. 
To calculate the loss of mixup networks, we first randomly select a mixing ratio $\lambda$ and a temperature prompt $P_2$ that differs from the given sample $P_1$. Using $f_{\phi_P}$, we generate a mixed prompt $P_{m}$. We then search for a molecule structure feature $S_2$, which is as similar to the given sample $S_1$ as possible, from the data pool of temperature $P_2$. We use MSE here to measure the similarity between two structure features. 
By leveraging $f_{\phi_{S}}$ to these two features, we obtain the mixed structure feature $S_{m}$. Also, we can calculate the mixed ground-truth $x^{GT}_{m}$ using the ground-truth positions of these two molecule structures, which we denote as $x^{GT}$ and $(x')^{GT}$ respectively. We can formulate this as:

\begin{equation}
\begin{aligned}
     T_{m} &= f_{\phi_{P}}(P_1,P_2,\lambda), \\
     S_{m} &= f_{\phi_{S}}(S_1,S_2,\lambda), \\
     x^{GT}_{m} &= (1-\lambda)x^{GT} + \lambda * (x')^{GT}.
\end{aligned}
\end{equation}

Finally, we input $P_{m}$ and $S_{m}$ to the pre-trained EGNN network to get a position $x_{m}$ and compute the MSE loss with  $x^{GT}_{m}$.

In the training process, we alternatively train mixup networks $f_\phi$ and the backbone network $f_\theta$, with the goal as follows:

\begin{equation}
    \min\limits_{\theta,\phi}L_{m}(x_{m},x^{GT}_{m})
\end{equation}

We decouple the training of parameters $\phi$ and $\theta$ by utilizing the technique of stopping the gradient. We denote this alternative training process as $L_m^\theta$ and $L_m^\phi$. The loss of our first-stage pre-training can be summarized into :

\begin{equation}
    L_{first} = L(x,x^{GT})+L_{m}^\theta(x_{m},x^{GT}_{m})+L_{m}^\phi(x_{m},x^{GT}_{m})
    \label{loss1}
\end{equation}

The aforementioned training schema of mixup networks serves to augment the training of our prompt-based protein simulation networks,  allowing for the synthesis of structure data and different temperatures. In this way, we can smoothly increase the training difficulty of our networks, not only shaping the latent space of prompts more continuous, but also enhancing  training of the pre-trained network.

\subsection{Prompt Meta-Training}
\label{second stage}
\begin{figure}[h]
    \centering
    \includegraphics[scale=0.8]{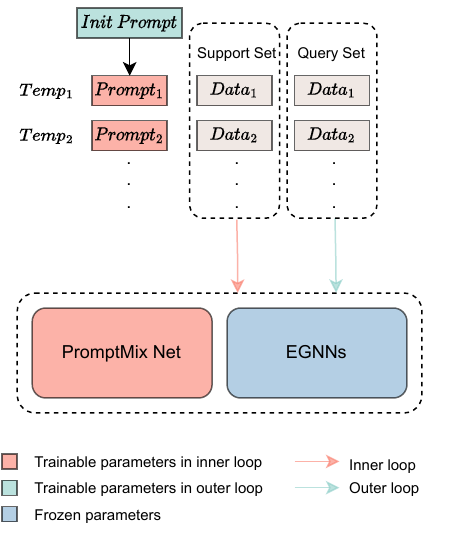}
    \caption{Illustration of meta training procedures. The network "EGNNs" stands for $\theta_O$, including prompt-agnostic encoder and the pre-trained network. "Prompt" refers to the prompt encoder. The StructMix network is not used in this process. All parameters are colored according to the stage they are in during the meta-training process. }
    \label{fig:meta}
\end{figure}

To enable prompt fine-tuning with minimal data and training iterations, we employ a few-shot meta-learning approach to the prompt and PromptMix while keeping other parts of the network frozen. The detailed illustration is shown in figure \ref{fig:meta}. In particular, we denote the parameters of the prompt as $\theta_P$ and the remaining parts as $\theta_O$, which consists of the prompt-agnostic encoder and the pre-training network EGNNs.  We separate our training dataset into support data and query data, which can be denoted as $D_{sup} = \{T^i_{sup}, x^i_{sup}, h^i_{sup}\}$ and $D_{que} = \{T^i_{que}, x^i_{que}, h^i_{que}\}$ respectively as in the traditional few-shot learning settings. In the first meta-training epoch, we initialize our PromptMix and prompt using the weights from first-stage pre-training.

\textbf{Inner Loop} In the inner loop, we train our PromptMix network $\phi_P$ and prompt $\theta_P$ using data $D_{sup}$. Similar to the first stage, we employ an alternative stop-gradient training procedure. Specifically, given a certain temperature prompt $P_i$, we update our PromptMix network $\phi_{P_i}$ and perform gradient descent on $\theta_{P_i}$, but without actually updating its init parameters in our model. Let $\alpha$ be the learning rate of the PromptMix network and $\beta$ be the learning rate of the prompts. The process of parameter updates can be described as follows:
\begin{equation}
    \theta_{P_i}' = \theta_{P_i} - \beta (\nabla_{\theta_P}L(f_{\theta_{P_i},\theta_O})+\nabla_{\theta_P}L^{\theta}_{m}(f_{\theta_{P_i},\theta_O,\phi_{P_i}}))
\end{equation}
\begin{equation}
    \phi_{P_i} = \phi_{P_i} - \alpha \nabla_{\phi_P}L^\phi_{m}(f_{\theta_{P_i}',\theta_O,\phi_{P_i}})
\end{equation}

\textbf{Outer Loop} In the outer loop, we use $\phi_P$ and $\theta_P'$ as model parameters and aim to find a better initialization point for the init prompt. Suppose we have $n$ temperature prompts in the inner loop, each can be written as $P_{i}$. Let $\gamma$ be the learning rate when updating parameters of the init prompt. Using the query dataset $D_{que}$ as inputs, the init parameters of prompt $\theta_P$ are updated as follows:

\begin{equation}
\begin{aligned}
    \theta_P = \theta_P - \gamma \nabla_{\theta_P}(\sum\limits_{P_{i=1}}^n (L(f_{\theta_{P_i}',\theta_O})+\\L^{\theta}_{m}(f_{\theta_{P_i}',\theta_O,\phi_{P_i}})))
    \end{aligned}
\end{equation}

\subsection{Testing Process}

Given the support data $D_{sup}$ under an unseen temperature, we adapt the parameters of prompt $\theta_P$ in the inner loop during the testing process. Specifically, we only rely on the MSE loss excluding any mix-up procedure, for fine-tuning the prompt parameters. \\
In the end, The refined prompt parameters are employed to estimate the atomic positions in subsequent datasets under the given temperature.
\begin{equation}
    \theta_{P}' = \theta_{P} - \beta \nabla_{\theta_P}L(f_{\theta_{P},\theta_O})
\end{equation}

\section{Experiments and Analysis}
\definecolor{lightteal1}{rgb}{0.9, 0.95, 1.0}
\definecolor{lightteal2}{rgb}{0.9, 1.0, 0.9}
\begin{table*}[!ht]
\renewcommand\arraystretch{1.3}
\centering
\begin{tabular}{cccccccc}
\hline
& & $Test_1$ & $Test_2$ & $Test_3$ & $Test_4$ & $Test_5$ & $Avg.$ \\
\hline
         & Training Temp. & 0.5585 & 0.6386 & 0.5945 & 0.5413 & 0.5841 & 0.5843 \\
\rowcolor{lightteal2}\cellcolor{white} Prompt Only & Unseen Temp.  & 0.5623 & 0.6506 & 0.5675 & 0.6357 & 0.6542 & 0.6141\\
\rowcolor{lightteal1}\cellcolor{white}          & OOD Temp.   & 1.2594 & 0.7102 & 0.6437 & 0.6445 & 1.2467 & 0.9009\\
\hline
            & Training Temp. & 0.5550 & 0.5616 & 0.5751 & 0.6049 & 0.5551 & 0.5703\\
\rowcolor{lightteal2}\cellcolor{white} Without Meta-Training & Unseen Temp.  & 0.5591 & 0.5581 & 0.5624 & 0.6071 & 0.5656 & 0.5705\\
\rowcolor{lightteal1}\cellcolor{white}             & OOD Temp.   & 0.6621 & 0.7046 & 0.6625 & 0.6589 & 0.6755 & 0.6732\\
\hline
                  & Training Temp. & 0.5555 & 0.5539 & 0.5547 & 0.5562 & 0.5567 & \bf{0.5554}\\
\rowcolor{lightteal2}\cellcolor{white} Ours              & Unseen Temp.  & 0.5544 & 0.5543 & 0.5568 & 0.5557 & 0.5558 & \bf{0.5554}\\
\rowcolor{lightteal1}\cellcolor{white}                   & OOD Temp.   & 0.6672 & 0.6650 & 0.6650 & 0.6615 & 0.6572 & \bf{0.6631}\\
\hline
\end{tabular}
\caption{Results of testing losses measured in terms of MSE loss under various temperature settings, with each test conducted after 3 epochs of fine-tuning. The loss of a temperature set is the average loss of each of the temperatures in this set. The testing process repeated for 5 times, and the average loss is reported.}
\label{result}
\end{table*}

In this session, we will demonstrate our purposed model in protein simulation task, to show the effect of generalization capabilities to unseen and out-of-distribution prompts with few-shot fine-tuning.

\subsection{Setup}

\textbf{Dataset} We primarily conduct Molecular Dynamics (MD) tasks involving the same protein molecule under different temperature conditions. Temperature plays a crucial role in influencing the properties and behavior of molecule structures. The activeness of molecules can be significantly affected by temperature variations. For instance, proteins demonstrate diverse thermal stability, with some maintaining their structure and function at high temperatures, while others may unfold or denature due to the disruption of weak non-covalent bonds holding the protein's structure together \citep{dong2018structural, julio2018thermal}. Conversely, at low temperatures, proteins can become less flexible and more rigid due to reduced activeness. Understanding and considering the impact of temperature on molecule structures is essential for accurate modeling and predictions in various scientific fields, including biochemistry and molecular biology.

Regarding the target protein, we have selected chignolin for our experiments. Chignolin is a peptide known to form a stable $\beta$-hairpin structure in water. Despite consisting of just 10 residues (GYDPETGTWG), it exhibits typical protein characteristics, folding into a distinctive structure and displaying a cooperative thermal transition between its unfolded and folded states \citep{honda200410}. Also, the structure of this protein is affected by temperature \citep{sumi2019theoretical}. With increasing temperature, the proportion of folded chignolin decreases, and it may adopt helical conformations and eventually unfold into extended structures. Hence, chignolin is not only faster to train but also retains the essential characteristics of proteins, making it a suitable representative for our experiments. To demonstrate our ability to generate new protein molecules with unseen prompts using limited data, we generate our own MD dataset using the openmm library \citep{eastman2017openmm}, a highly efficient toolkit widely employed for molecular simulations.

This dataset consists of data points from the same protein molecule (chignolin) but under varying temperatures. Specifically, we utilize 10,000 step data points for the two-stage pre-training, respectively obtained at \texttt{Training Temperature Set} $T_{train}=\{280K, 300K, 320K\}$. 
Additionally, we use 3,000 data points for testing, respectively obtained at temperatures of 280K, 285K, 290K, 295K, 300K, 305K, 310K, 315K, 320K, 350K. Notably, $T_{unseen}=\{285K, 290K, 295K, 305K, 310K, 315K\}$ represent for \texttt{Unseen Temperature Set}, while $T_{OOD}=\{350K\}$ is \texttt{Out-of-Distribution Temperature Set}.\\

\textbf{Implementation}  In the pre-training stage, we use LION \citep{chen2023symbolic} as our optimizer with learning rate 1e-4 and set the weight decay to 1e-5.  As for the hyperparameters setting of curriculum learning, we set $N_{period} = 256$, $N_{min} = 16$, $E_{period} = 10, E_{pre} =50$.\\

In the meta-training stage, we generally follow the setting of MAML \citep{finn2017modelagnostic}, except that we fine-tune on parameters of prompt and PromptMix in the inner loop, while training the initial prompt parameters in the outer loop, as described in our Methodology session. We use Adam \citep{kingma2017adam} as our optimizer, and set the learning rate $\alpha = \gamma = 0.001$, and the learning rate in the inner loop $\beta = 0.01$.

\subsection{Baselines and Evaluation Metrics}

We perform experiments in three distinct settings.

\begin{itemize}
    \item \textbf{Prompt Only}: This baseline excludes the mixup network and the meta-training process. We set up this baseline to show the effectiveness of our mixup network.

    \item \textbf{Without Meta-Training}: This baseline adds the mixing part but excludes the meta-training process.

    \item \textbf{Ours}: This is our final version with all the components including mixup network and meta-training process. 
\end{itemize}

We evaluate the performance of our design by the MSE loss during testing. Specifically, we fine-tune the prompt encoder for three epochs under various temperature conditions. For this fine-tuning process, we randomly choose half of the available test data under each temperature to form the support set $D_{sup}$, while the remaining half is used to evaluate the testing losses. Since the fine-tuning process might have some randomness, we repeat it five times. 
The quantitive result is shown in Table \ref{result}. \\

\subsection{Results and Analysis}

In comparison with \textbf{Prompt only} and \textbf{Without Meta-Training}, we can observe that the testing results of the latter one are close to the first one under $T_{train}$, slightly better under $T_{unseen}$, and significantly outperform the first one under $T_{OOD}$. 
This demonstrates that by incorporating the mixup network within the curriculum learning schema, our method
greatly enhances the generalization ability of the networks with the same data scale by shaping the distribution of prompt vectors in a continuous manner.
Moreover, our mixup networks are just simple linear networks, and thus our method can work well without largely increasing the model scale. \\

Although \textbf{Without Meta-Training} performs well in most of the cases, we still observe some instabilities with more trials. This could be caused by the random initialization of prompt parameters and different support sets under the same temperature. To solve this problem, we add the meta-learning process to give better fixed initial prompt parameters so as to further shape the final distribution of the fine-tuned prompts. As shown in Table \ref{result}, \textbf{Ours} method performs no outlier under all temperatures. \\

\begin{figure}[hb]
    \centering
    \includegraphics[scale = 0.65]{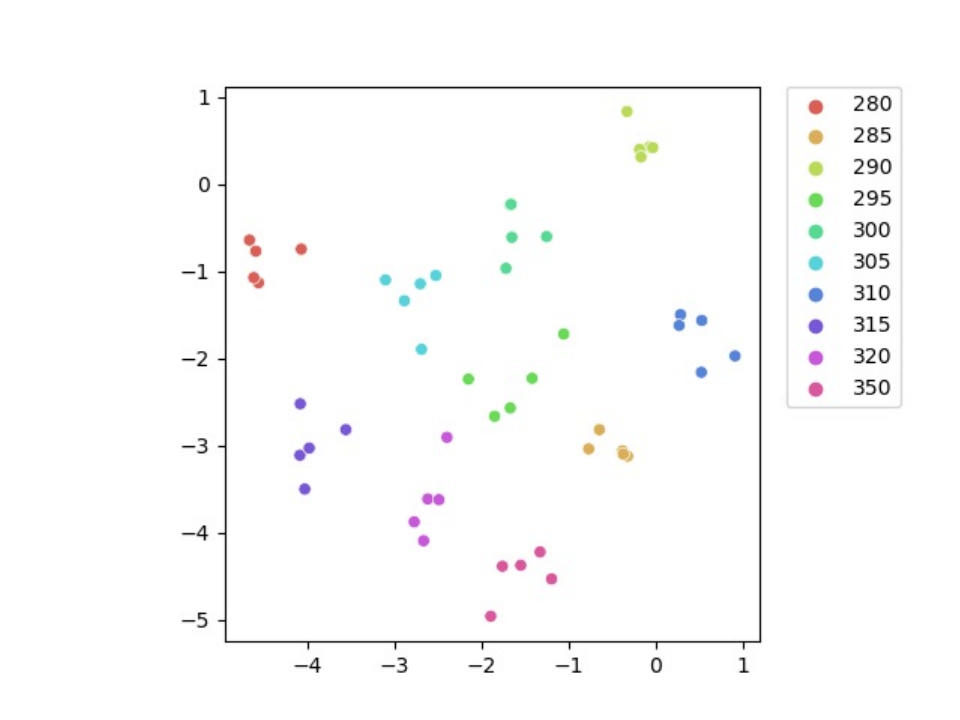}
    \caption{Visualization of Prompt Distribution}
    \label{fig:tsne}
\end{figure}

It should be noted that our model is only pre-trained on three distinct temperature settings in $T_{train}$ but can generalize to seven unseen temperature settings in $T_{unseen}$ and out-of-distribution settings in $T_{OOD}$. 
This substantial gap between training and testing demonstrates the remarkable generalization capability and the continuous latent space of our model. Additionally, during the fine-tuning process, only the prompt encoder, which is a linear network, is modified. As a result, the fine-tuning process is efficient regarding samples and computational resources.

Furthermore, we use t-SNE \citep{van2008visualizing} to visualize the distribution of prompts after fine-tuning. 
Figure \ref{fig:tsne} shows that prompts of different temperatures are highly separable. This observation indicates that the prompt space has been effectively shaped. In other words, the prompt encoder is capable of accurately recognizing both unseen and out-of-distribution (OOD) prompts, showcasing its robustness and generalization capability to various temperature conditions.

\section{Conclusion}
In this paper, we propose a two-stage pre-training method for temperature-based molecule generation tasks. By incorporating mixup data augmentation techniques and meta-training, we enhance the sample-efficiency of the fine-tuning process and improve the accuracy of molecule structure prediction for unseen and out-of-distribution (OOD) prompts, even with limited data and shorter fine-tuning epochs. This prompt-tuning framework can be further extended to other prompt-based tuning networks for future developments.

\bibliography{custom}
\bibliographystyle{apalike}
\end{document}